\documentclass[letterpaper, 10 pt, conference]{ieeeconf}  

\IEEEoverridecommandlockouts                              

\usepackage{graphics} 
\usepackage{epsfig} 
\usepackage{amsmath} 
\usepackage{multirow}

\title{\LARGE \bf
3D Shape-Based Myocardial Infarction Prediction Using Point Cloud Classification Networks
}

\author{Marcel Beetz$^{1}$$^{\star}$$^{\dagger}$, Yilong Yang$^{1}$$^{\star}$, Abhirup Banerjee$^{2}$$^{\dagger}$, Lei Li$^{1}$, Vicente Grau$^{1}$
\thanks{$^{\star}$ Authors contributed equally}%
\thanks{$^{\dagger}$ Corresponding author}%
\thanks{$^{1}$M. Beetz, Y. Yang, L. Li, and V. Grau are with the Institute of Biomedical Engineering, Department of Engineering Science, University of Oxford, Oxford OX3 7DQ, UK
        ({\tt\small marcel.beetz@eng.ox.ac.uk}, {\tt\small yilong.yang@mansfield.ox.ac.uk}, {\tt\small lei.li@eng.ox.ac.uk}, {\tt\small vicente.grau@eng.ox.ac.uk})}%
\thanks{$^{2}$A. Banerjee is with the Institute of Biomedical Engineering, Department of Engineering Science, University of Oxford, Oxford OX3 7DQ, UK and also with the Division of Cardiovascular Medicine, Radcliffe Department of Medicine, University of Oxford, Oxford OX3 9DU, UK
        {\tt\small abhirup.banerjee@eng.ox.ac.uk}}%
}

\begin{document}
\bstctlcite{IEEEexample:BSTcontrol}

\maketitle
\thispagestyle{empty}
\pagestyle{empty}

\begin{abstract}
Myocardial infarction (MI) is one of the most prevalent cardiovascular diseases with associated clinical decision-making typically based on single-valued imaging biomarkers. However, such metrics only approximate the complex 3D structure and physiology of the heart and hence hinder a better understanding and prediction of MI outcomes. In this work, we investigate the utility of complete 3D cardiac shapes in the form of point clouds for an improved detection of MI events. To this end, we propose a fully automatic multi-step pipeline consisting of a 3D cardiac surface reconstruction step followed by a point cloud classification network. Our method utilizes recent advances in geometric deep learning on point clouds to enable direct and efficient multi-scale learning on high-resolution surface models of the cardiac anatomy. We evaluate our approach on 1068 UK Biobank subjects for the tasks of prevalent MI detection and incident MI prediction and find improvements of $\sim$13\% and $\sim$5\% respectively over clinical benchmarks. Furthermore, we analyze the role of each ventricle and cardiac phase for 3D shape-based MI detection and conduct a visual analysis of the morphological and physiological patterns typically associated with MI outcomes.
\newline

\indent \textit{Clinical relevance}—  The presented approach enables the fast and fully automatic pathology-specific analysis of full 3D cardiac shapes. It can thus be employed as a real-time diagnostic tool in clinical practice to discover and visualize more intricate biomarkers than currently used single-valued metrics and improve predictive accuracy of myocardial infarction.
\newline
\end{abstract}


\section{INTRODUCTION}
\label{sec:intro}

Myocardial infarction (MI) is a common manifestation of coronary artery disease, the deadliest pathology in the world \cite{khan2020global}. In current clinical practice, its diagnosis and treatment typically involve the acquisition of cardiac cine magnetic resonance (MR) images as the gold standard imaging modality for cardiac anatomy and function assessments \cite{reindl2020role}. However, current clinical decision-making is often guided by single-valued biomarkers, such as ejection fraction, which are directly calculated from 2D MR imaging (MRI) slices to evaluate cardiac anatomy and physiology on a purely global level \cite{reindl2020role}. Consequently, considerable research efforts have focused on developing methods that can take more comprehensive image information into account \cite{bernard2018deep,isensee2018automatic,wolterink2018automatic,zhang2019deep,li2023myops}. However, all these approaches still only approximate the true 3D structure of the heart based on 2D images or image-derived features and therefore neglect more complex and localized changes in 3D cardiac shapes, which play a crucial role in improving the understanding, prediction, and management of MI outcomes \cite{suinesiaputra2017statistical,corral2022understanding,beetz2022interpretable,beetz2023post,beetz2023mesh}.

In this work, we investigate the utility of full 3D cardiac shape representations in the form of point clouds for the detection and prediction of MI events. To this end, we propose a novel fully-automatic MI detection pipeline, which first reconstructs 3D cardiac anatomy point clouds from raw cine MR images and then employs targeted point cloud networks for the MI classification task. The network architectures of its individual components are based on recent advances in point cloud-based deep learning to enable efficient multi-scale feature learning directly on anatomical surface data. Deep learning approaches for point cloud data have recently been increasingly used in the field of cardiac image analysis for a variety of applications, such as 3D surface reconstruction \cite{zhou2019one,ye2020pc,beetz2021biventricular,chen2021shape,beetz2023point2mesh}, image segmentation \cite{ye2020pc,chang2020automatic}, pathology classification \cite{chang2020automatic}, or 3D anatomy and function modeling \cite{beetz2021generating,beetz2021predicting,beetz2022combined,beetz2022multi,li2023deep}. In this work, we specifically study 3D anatomical representations of the left and right ventricles at both ends of the cardiac cycle and their effect on prior and future MI. To the best of our knowledge, this is the first MRI-based point cloud deep learning approach to focus on MI prediction directly from 3D cardiac shapes.

\section{METHODS}
\label{sec:methods}

\begin{figure*}[!ht]
\centerline{\includegraphics[width=\textwidth]{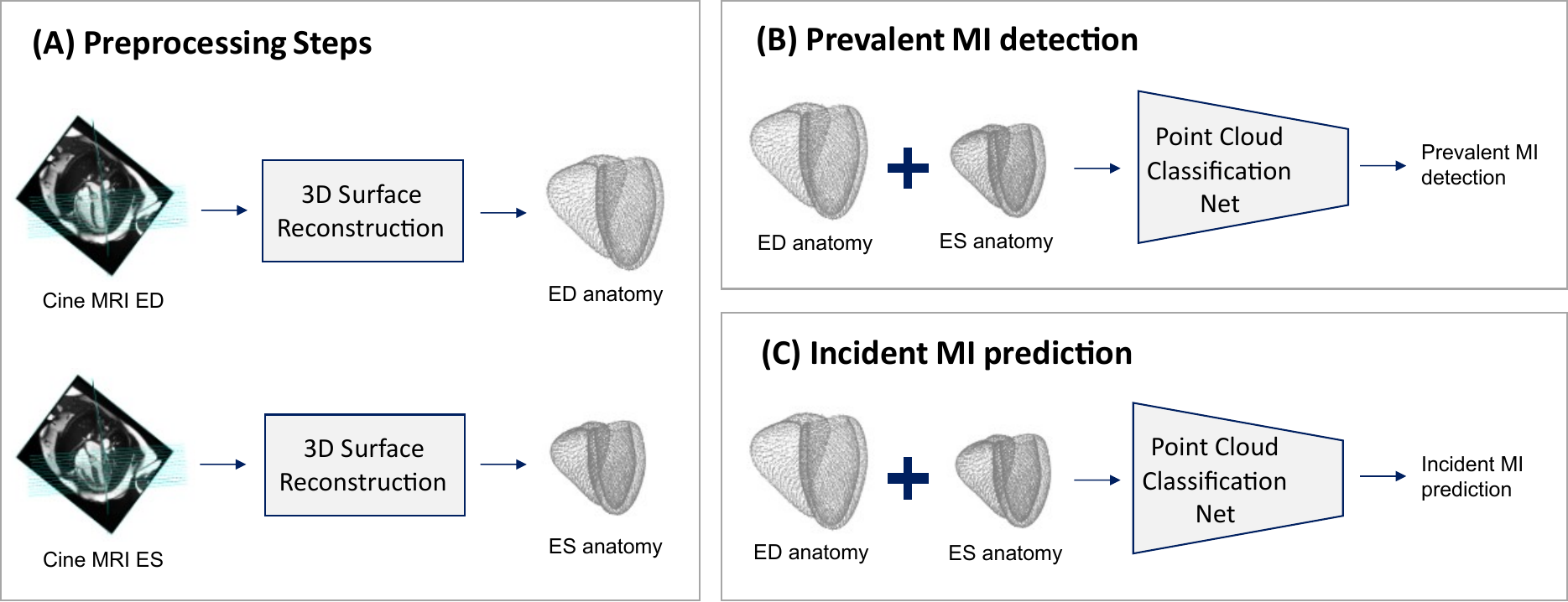}}
\caption{Overview of our proposed 3D anatomy-based infarction detection pipeline. We first reconstruct 3D point cloud representations of the cardiac anatomy from corresponding cine MRI acquisitions at the ED and ES phases of the cardiac cycle using a fully automatic multi-step pipeline (A) \cite{beetz2021biventricular}. We then directly input the obtained 3D anatomies into separate specialized point cloud classification networks for the tasks of prevalent MI detection (B) and incident MI prediction (C).}
\label{fig:overview}
\end{figure*}

\subsection{Dataset}
\label{ssec:dataset}

Our dataset consists of 1068 subjects of the UK Biobank study for which cine MR images were acquired using a balanced steady-state free precession (b-SSFP) protocol \cite{petersen2015uk}. An MI event after the imaging date (incident MI) was recorded for 235 subjects, while 294 subjects suffered an MI event prior to imaging (prevalent MI). The remaining 539 subjects were selected to be free of any diseases associated with the cardiovascular system and are used as normal control cases for our analysis. We follow the disease definition as proposed in previous work \cite{bai2020population} and use the UK Biobank field ID 42,000 to identify both incident and prevalent MI subjects.

\subsection{Infarction Detection Pipeline}
\label{ssec:pipeline_overview}

Our proposed 3D anatomy-based MI detection pipeline consists of multiple fully automatic steps as illustrated in Fig.~\ref{fig:overview}.
It receives the cine MRI acquisitions at both the end-diastolic (ED) and end-systolic (ES) phases of the cardiac cycle as inputs. Based on these inputs, we reconstruct the corresponding 3D biventricular anatomy models at both phases using a multi-step cardiac surface reconstruction approach \cite{beetz2021biventricular} (Fig.~\ref{fig:overview}-A). It first segments the left ventricular (LV) endocardium, LV epicardium, and right ventricular (RV) endocardium in the short-axis and four-chamber long-axis (LAX) slices of the MRI acquisition with separate pre-trained fully convolutional neural networks \cite{bai2018automated} and in the two-chamber LAX images using a U-Net with adversarial training. The resulting segmentation contours of all image slices are then placed into 3D space as sparse point clouds \cite{banerjee2021ptrsa} before a point cloud completion network is employed to correct the motion-induced slice misalignment and output dense point cloud representations of the 3D cardiac anatomy.

These 3D cardiac anatomies are then used as inputs to point cloud classification networks for the tasks of prevalent MI classification (Fig.~\ref{fig:overview}-B) and incident MI prediction (Fig.~\ref{fig:overview}-C). For each task, we study both ES only and combined ED and ES anatomy inputs as implicit and explicit representations of 3D shape-based cardiac function. In the latter case, we concatenate the ED and ES point clouds before feeding them into the point cloud classification network, giving it direct access to all available anatomical information at both phases. In addition, we investigate the utility of the RV as part of a biventricular representation of 3D cardiac shape for MI, by using first only LV anatomies, and then combined LV and RV anatomies as network inputs. We analyze the effect of these two different shape inputs for both MI classification tasks and for each of the two temporal input types, resulting in a total of 8 different experiments.

\subsection{Point Cloud Classification Network}
\label{ssec:point_cloud_network}

We choose PointNet \cite{qi2017pointnet} as the architectural basis of our point cloud classification network and adjust it for the task of binary MI classification of 3D point cloud representations of cardiac anatomy and function. To this end, we first use a sigmoid activation layer at the end of PointNet's classification branch to obtain binary prediction probabilities as the network's output. We then tune the drop-out probabilities in the last multi-layer perceptron part of the network based on a grid search procedure. We train our network using the binary cross entropy loss and the Adam optimizer with a mini-batch size of 20 and a learning rate of 1E-6 for 200 epochs on an RTX 2060S GPU with 8~GB memory.

\section{EXPERIMENTS AND RESULTS}
\label{sec:experiments}

\subsection{Prevalent Infarction Detection}
\label{ssec:prevalent_mi}
In our first experiment, we assess whether the high-resolution 3D point cloud representations of the cardiac anatomy contain more information about prevalent MI events than corresponding global clinical benchmarks and whether a point cloud-based deep learning network is able to successfully extract them without any manual intervention. Furthermore, we analyze the importance of different cardiac substructures and cardiac phases for this task. To this end, we train four separate point cloud classification networks using the ES LV anatomy, the combined ED and ES LV anatomies, the ES biventricular anatomy, and the combined ED and ES biventricular anatomies as inputs to the respective networks. We then select widely used clinical metrics (ES volume, ejection fraction) for the LV and RV as our comparative benchmarks and input them as independent variables in four separate logistic regression models, each trained on the same dataset and task as the corresponding point cloud networks. We conduct a four-fold cross validation experiment in each case and report the results in terms of area under the receiver operating characteristic (AUROC) scores in Table~\ref{tab:prevalent_mi_detection}.

\begin{table}[!h]
    \caption{Results of the proposed and benchmark methods for prevalent MI detection}
    \def\arraystretch{1.5}\tabcolsep=3pt
    \centering
    \begin{tabular}{p{40pt}p{75pt}p{50pt}p{40pt}}
        \hline
        Anatomy & Input & Method & AUROC  \\
        \hline
        \multirow{4}*{LV} &  ES Volume & Regression  &  0.654  \\
            &   ES 3D Shape &  Proposed & \textbf{0.705}  \\
        \cline{2-4}
        & Ejection Fraction & Regression  &  0.670  \\
            &   ED+ES 3D Shape & Proposed   & \textbf{0.725}   \\
        \hline         
        \multirow{4}*{LV+RV} & ES Volume & Regression &  0.641  \\
             &  ES 3D Shape & Proposed  & \textbf{0.699}  \\
        \cline{2-4}
        & Ejection Fraction & Regression  &  0.671  \\
            &  ED+ES 3D Shape & Proposed  & \textbf{0.758}  \\
                    
        \hline
    \end{tabular}
    \label{tab:prevalent_mi_detection}
\end{table}

We find that 3D shape-based point cloud classification networks outperform the respective clinical benchmarks for all cardiac phases and anatomical substructures with an average relative difference of $\sim$10\% in terms of AUROC. As expected, the combined biventricular input at ED and ES achieves the highest score and a $\sim$13\% outperformance of its respective clinical benchmark.

\begin{figure}[htb]
	\begin{minipage}[b]{1.0\linewidth}
		\centering
		\centerline{\includegraphics[width=8.2cm]{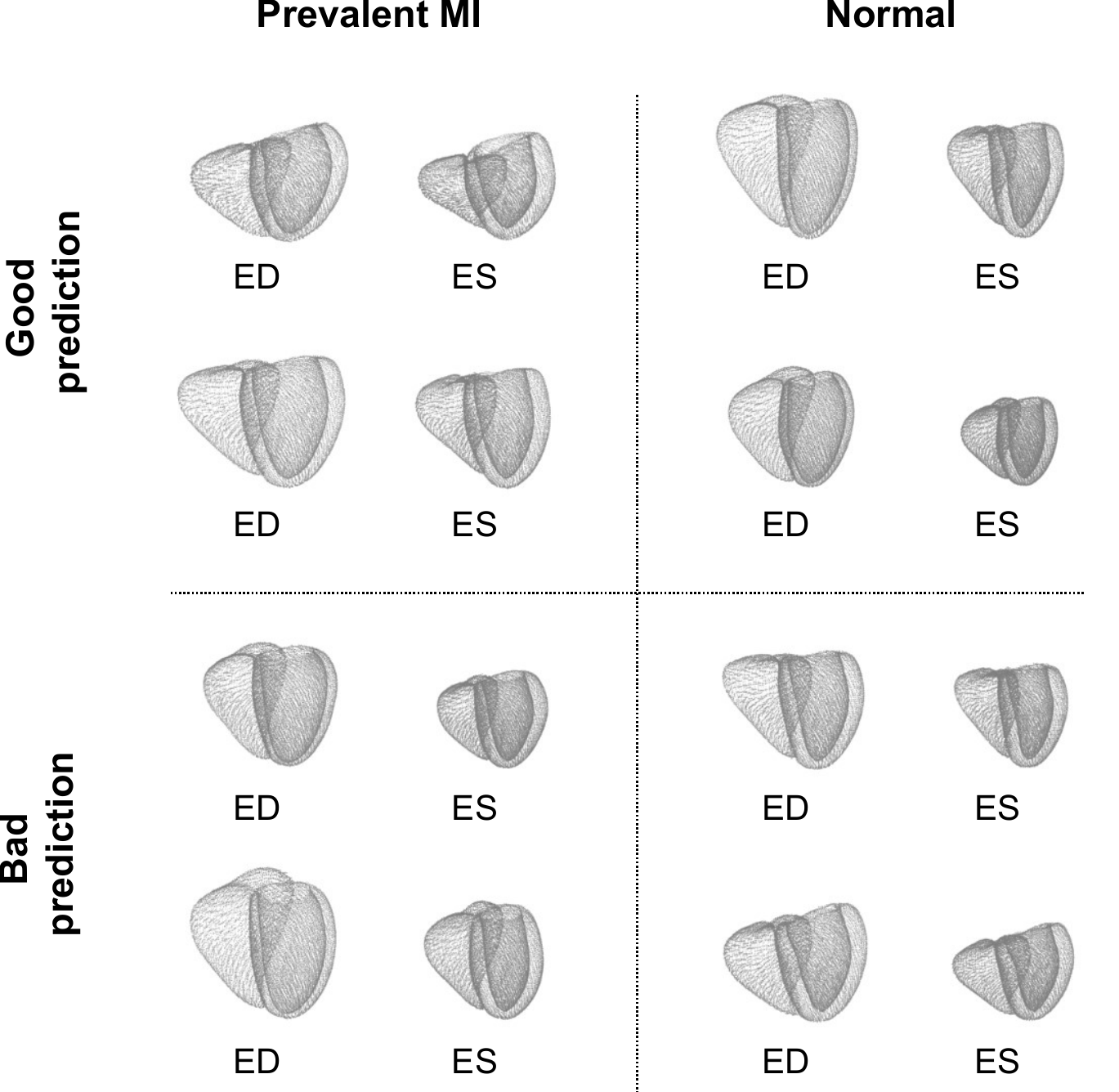}}
	\end{minipage}
	\caption{Sample anatomies that resulted in good and bad predictions (rows) for prevalent MI and normal cases (columns).}
	\label{fig:prevalent_detection_qualitative}
\end{figure}

Following this quantitative evaluation, we further investigate which 3D anatomical shape features are typically associated with prevalent MI cases by the network. To this end, we select two representative cases corresponding to good and poor network predictions on the test dataset for both MI and normal cases and visualize them in Fig.~\ref{fig:prevalent_detection_qualitative}.

We observe that good network predictions for prevalent MI subjects are more likely to occur in cases of reduced myocardial thickening and smaller overall volume changes between the ED and ES phases, and vice versa for normal cases. Bad predictions more commonly happen when these associations are weakened or reversed.

\subsection{Incident Infarction Prediction}
\label{ssec:incident_mi}

In addition to detecting prevalent MI events, we investigate whether 3D anatomy-based patterns learned by point cloud networks are also beneficial for the prediction of incident MI events. We follow a similar experimental setup as for prevalent MI classification in Sec.~\ref{ssec:prevalent_mi} and train four separate networks and their corresponding clinical regression benchmarks for the binary prediction of incident MI events using four-fold cross validation and the AUROC evaluation metric (Table~\ref{tab:incident_mi_prediction}). We again use the full 3D shapes (LV ES, LV ED+ES, LV+RV ES, LV+RV ED+ES) as neural network inputs and the respective clinical metrics (LV ES volume, LV ejection fraction, LV+RV ES volume, LV+RV ejection fraction) as independent regression variables.

\begin{table}[!h]
    \caption{Results of the proposed and benchmark methods for incident MI prediction}
    \def\arraystretch{1.5}\tabcolsep=3pt
    \centering
    \begin{tabular}{p{40pt}p{75pt}p{50pt}p{40pt}}
        \hline
        Anatomy & Input & Method & AUROC  \\
        \hline
        \multirow{4}*{LV} &  ES Volume & Regression  &  0.632  \\
            &   ES 3D Shape &  Proposed & \textbf{0.660}  \\
        \cline{2-4}
        & Ejection Fraction & Regression  &  0.635  \\
            &   ED+ES 3D Shape & Proposed   & \textbf{0.654}   \\
        \hline         
        \multirow{4}*{LV+RV} & ES Volume & Regression &  0.620  \\
             &  ES 3D Shape & Proposed  & \textbf{0.651}  \\
        \cline{2-4}
        & Ejection Fraction & Regression  &  0.618  \\
            &  ED+ES 3D Shape & Proposed  & \textbf{0.646}  \\
                    
        \hline
    \end{tabular}
    \label{tab:incident_mi_prediction}
\end{table}

We find that the 3D shape-based point cloud network is able to outperform the respective clinical benchmark for both cardiac phases and ventricles by $\sim$4\% on average. The best score is achieved by the combined ventricular anatomy at ES with a $\sim$5\% improvement. When visually examining the results, we observe similar patterns as in our prevalent MI detection experiments (Sec.~\ref{ssec:prevalent_mi}) with a generally higher probability of accurate MI prediction for smaller changes in myocardial thickness between ED and ES phases.

\section{DISCUSSION AND CONCLUSION}
\label{sec:discussion}

We have presented a novel end-to-end point cloud-based deep learning pipeline for the detection of both prior and future MI events based on 3D cardiac shapes. In our experiments, the method has been able to outperform corresponding clinical benchmarks for both classification tasks using a variety of different inputs. On the one hand, this indicates that full 3D cardiac shapes contain more infarction-related information than current single-valued clinical biomarkers, which is in line with previous works \cite{suinesiaputra2017statistical,corral2022understanding} and promises to improve both patient risk stratification and the implementation of preventive measures. On the other hand, it shows that the architectural design of our pipeline is adequately chosen to successfully extract relevant biomarkers directly from the 3D anatomical shapes. Hereby, the selected point cloud representation of cardiac surface data considerably reduces the memory requirements compared to previous voxelgrid-based approaches. Combined with the fully automatic pipeline design, this allows for faster execution speeds, wider applicability, and easy scaling to both higher resolutions and large numbers of patients in real time. 

In our experiments, we also find better predictive performance for prevalent compared to incident MI cases. We hypothesize that this is primarily caused by the more easily visible morphological changes of post-MI cardiac remodeling, which the network is able to capture. While the addition of RV information achieved mixed results, the inclusion of anatomies at both ED and ES phases generally improved predictive accuracy, which corroborates previous findings on the importance of 3D LV contraction information for MI detection \cite{suinesiaputra2017statistical,corral2022understanding}. While we focused on the role of 3D shapes in this study, we believe that the pipeline can be easily extended to include other patient-specific information with a potential to further improve the understanding of MI events.

\addtolength{\textheight}{-12cm}   



\section*{ACKNOWLEDGMENT}
This research has been conducted using the UK Biobank Resource under Application Number ‘40161’. The authors express no conflict of interest.
The work of M. Beetz was supported by the Stiftung der Deutschen Wirtschaft (Foundation of German Business).
A. Banerjee is a Royal Society University Research Fellow and is supported by the Royal Society Grant No. URF{\textbackslash}R1{\textbackslash}221314.
The work of A. Banerjee was partially supported by the British Heart Foundation (BHF) Project under Grant PG/20/21/35082.
The works of V. Grau and L. Li were supported by the CompBioMed 2 Centre of Excellence in Computational Biomedicine (European Commission Horizon 2020 research and innovation programme, grant agreement No. 823712).
L. Li was partially supported by the SJTU 2021 Outstanding Doctoral Graduate Development Scholarship.


\bibliographystyle{IEEEtran}
\bibliography{refs}

\end{document}